\documentclass[conference]{IEEEtran}
\IEEEoverridecommandlockouts

\usepackage{cite}
\usepackage{amsmath,amssymb,amsfonts}
\usepackage{algorithmic}
\usepackage{graphicx}
\usepackage{textcomp}
\usepackage{xcolor}
\usepackage{booktabs}
\usepackage{makecell}
\usepackage[hidelinks]{hyperref}
\usepackage{csvsimple-legacy}
\usepackage{url}
\def\BibTeX{{\rm B\kern-.05em{\sc i\kern-.025em b}\kern-.08em
    T\kern-.1667em\lower.7ex\hbox{E}\kern-.125emX}}
\begin{document}

\title{
 \footnotesize
 \framebox[1.01\width]{\parbox{\dimexpr\linewidth-2\fboxsep-2\fboxrule}{If you cite this paper, please use this reference: N. Duboisset et al. PolyChirp: Multi-Species Birdsong Classification Using TinyML on Low-Power Acoustic Sensors. in Proceedings of the IEEE International Symposium on the Internet of Sounds (IS2). Cannes, France, October 2026.}}
  \ \\ \ \\ \ \\
 \huge
PolyChirp: Multi-Species Birdsong Classification Using TinyML on Low-Power Acoustic Sensors
}

\author{\IEEEauthorblockN{
Nathan Duboisset}
\IEEEauthorblockA{\textit{École Polytechnique}, Palaiseau, France \\
\textit{Freie Universität Berlin}, Berlin, Germany \\
nathan.duboisset@polytechnique.edu}
\and
\IEEEauthorblockN{
Zhaolan Huang}
\IEEEauthorblockA{\textit{Freie Universität Berlin} \\
Berlin, Germany \\
zhaolan.huang@fu-berlin.de}
\and
\IEEEauthorblockN{
Felix Bießmann}
\IEEEauthorblockA{\textit{Berlin University of Applied Sciences (BHT)} \\
Berlin, Germany \\
felix.biessmann@bht-berlin.de}
\and
\IEEEauthorblockN{
Roudy Dagher}
\IEEEauthorblockA{\textit{School of Engineering, HES-SO Valais-Wallis}\\Sion, Switzerland\\
roudy.dagher@hevs.ch}
\and
\IEEEauthorblockN{
Antoine Lavandier}
\IEEEauthorblockA{\textit{Inria} \\
France \\
antoine.lavandier@inria.fr}
\and
\IEEEauthorblockN{
Emmanuel Baccelli}
\IEEEauthorblockA{\textit{Inria}, France \\
\textit{Freie Universität Berlin}, Germany \\
emmanuel.baccelli@inria.fr}
}


\maketitle

\begin{abstract}
Recent progress in the field of TinyML has demonstrated that low-power hardware based on microcontrollers can achieve bird species monitoring in real time based on acoustic sensor data for an entire breeding period on a single battery charge. However, the state of the art on low-power microcontrollers was so far limited to binary classification of a single species. In contrast, real fauna monitoring deployments often target multiple species simultaneously. To address this challenge we develop PolyChirp, an approach combining biological domain expertise, automated dataset curation, neural architecture optimization and novel hardware to achieve multiclass bird species detection in the wild. PolyChirp is based on newly designed tiny multiclass models that leverage recent microcontrollers and hardware acceleration with a neural processing unit (NPU). 
We evaluate the predictive performance of these models, and we measure their computational performance -- flash footprint, latency, energy consumption -- on common microcontroller hardware. Our results demonstrate that PolyChirp matches or exceeds the TinyChirp architectures retrained under our protocol on single-species detection, and further achieves robust classification of up to 10 species simultaneously (macro $F_2$ up to $0.97$), while still fitting the flash, latency and energy budget of a low-power microcontroller sensor. A data-driven front-end redesign additionally makes on-device mel feature extraction $7$ to $11\times$ cheaper.
\end{abstract}

\section{Introduction}

Passive acoustic monitoring of birds usually means deploying battery-powered sensors in the field as autonomous recording units (ARUs) for a full breeding season, recording continuously. Months later, the devices and their filled memory cards are collected for analysis. Yet only a small fraction of these recordings is of interest. While a campaign targets one or more bird species, recordings end up also including in vast amounts sounds of wind, traffic, other animals etc. Crucially, storing spurious audio depletes the two budgets that constrain an ARU: memory and battery capacities. 

One approach to circumvent these bottlenecks is to use TinyML to decide quickly directly on the sensor whether the current acoustic signal is relevant or not.
The difficulty is that the sensor is a low-power microcontroller with a few hundred kilobytes of RAM~\cite{rfc7228}. Hence, the classifier has to be very small, and in particular, a big model such as BirdNET~\cite{birdnet} does not fit in memory.
In contrast, prior work such as TinyChirp~\cite{tinychirp} demonstrates how microcontroller-based hardware can detect a single target bird species in the field, accurately enough to discard most spurious audio and significantly improve these bottlenecks. 

However, such prior work still suffers from limitations in practice. 
The first is the number of species: a deployment usually follows several species at once, and a single-output detector can neither separate them nor report which one it heard. The second is portability across sites: the species that matter differ from one location to the next, so a model and a dataset built for one species in one place does not carry over, and redoing that data work by hand for every new site is exactly the cost one wants to avoid.

This paper thus introduces PolyChirp, a new multi-species on-device classification mechanism that can run on diverse microcontroller hardware. PolyChirp also provides an automated dataset generation scheme which facilitates re-training for different geographical locations and, accordingly, for different species subsets. In more detail, our main contributions are as follows:



\begin{itemize}
    \item We provide a geographical site-driven dataset builder that turns a location, a radius, and a target species count into a deployment-specific multi-class dataset. 
    \item We design and implement a family of multi-class TinyML models. 
    We publish the source code of our implementations.
    \item We provide a comparative performance evaluation of predictive performance of PolyChirp using different TinyML models, as well as measurements of Flash, inference latency, and energy consumption 
    on common microcontroller hardware with and without NPU acceleration. 
    \item We show that, retrained under our protocol, PolyChirp matches or exceeds the performance of previous work in the single-species case, while enabling use cases handling up to 10 species at once at macro $F_2$ up to $0.97$.
\end{itemize}

\section{Datasets}
\label{sec:datasets}

Most bird-audio benchmarks use a fixed, global species list. 
Instead, we automate dataset building based on the place where the device will be deployed. 

\label{sec:datasets:build}

\textbf{Dataset Creation Automation} -- To generate the dataset, we query Xeno-Canto~\cite{xenocanto} for every recording inside a bounding box, defined by coordinate, radius, and a target species count $N$. We run BirdNET~\cite{birdnet} on a random subsample of those recordings to catch species the metadata misses, and rank species by occurrence. By default, the first $N$ species become the target classes. The remaining species are pooled into a single \texttt{non\_target} class together with AudioSet~\cite{audioset} clips (speech, traffic, weather, machinery, indoor noise, etc.) and no-bird windows mined from recordings from the same area. A second primary source, the Macaulay Library~\cite{macaulay}, is queried in parallel to grow the target counts, using the same query system, to add more diverse recordings, instruments, and locations.

As an example, we run the procedure once for Paris with a 50\,km radius and $N=10$. The ten target classes range from 4870 to 10\,000 clips. The \texttt{non\_target} class holds 18\,800 clips, split roughly evenly between 8400 other birds, 8400 AudioSet clips and 2000 no-bird clips. Recordings from Xeno-Canto and Macaulay are MP3 formats at mixed sample rates (32k to 48\,kHz) and bit-rates (96k to 320\,kbps); AudioSet clips are 10\,s YouTube segments. We split each class 70/15/15 into train/val/test, stratified at the recording level so clips from the same source recording stay in the same split, to avoid data leakage.

\label{sec:datasets:preproc}

\textbf{Feature Selection and Pre-processing --} Every recording is decoded and resampled to 16\,kHz mono; levels are left as recorded. BirdNET then labels it: detections at confidence $\geq 0.92$ produce 3\,s clips around the detection, and recordings with no detection at all contribute a single random 3\,s window to the no-bird pool. Labelling with BirdNET instead of the metadata tag lets us use the whole length of an XC or Macaulay recording while keeping the per-clip class confidence high, which matters because community recordings often consist mostly of background noise.

\begin{figure}[t]
\centering
\includegraphics[width=\linewidth]{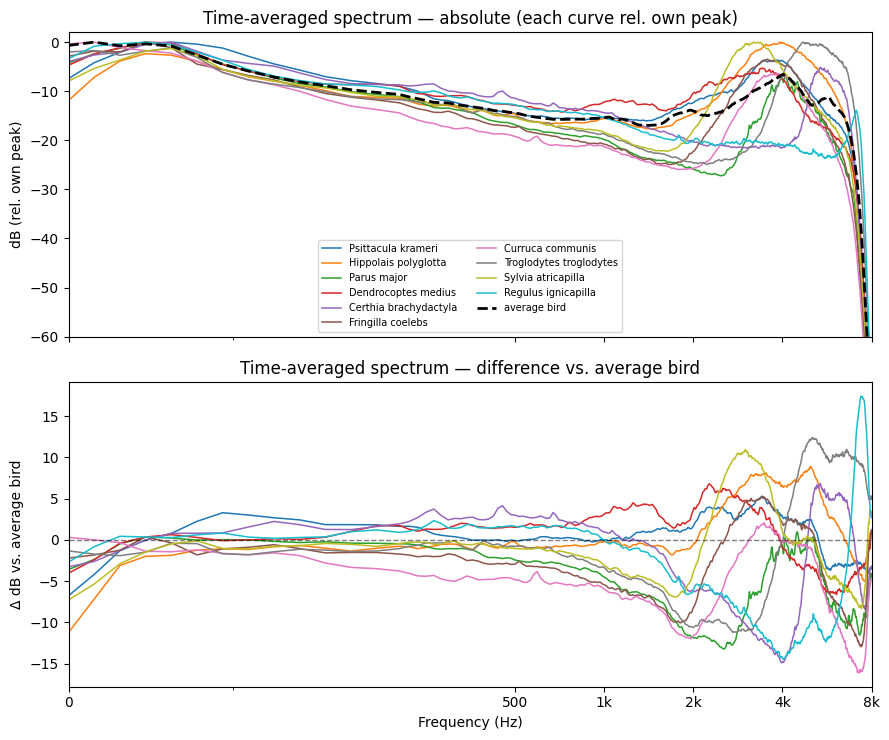}
\caption{Time-averaged spectrum of each \texttt{paris\_10} target species (each curve normalised to its own peak) against the pooled non-target ``average bird''. Top: absolute; bottom: difference vs.\ the average bird. Per-species separation is concentrated above 2\,kHz, motivating the \texttt{light\_mel} 2500--8000\,Hz variant.}
\label{fig:birds-spectrum}
\end{figure}

For the mel-spectrogram models we use two front-ends, \texttt{tinychirp\_mel} and \texttt{light\_mel}. The first, following TinyChirp~\cite{tinychirp}, is a 1024-point Hann window with a 256-sample hop (16\,ms) and 80 mel bins over 80--8000\,Hz, producing a $184\times 80$ log-mel spectrogram per 3\,s clip. The lighter \texttt{light\_mel} front-end shrinks every axis at once: a 512-point window with a 480-sample hop and 40 mel bins over 2500--8000\,Hz, producing $99\times 40$ spectrograms. 
The narrower band comes directly from the data: averaging the spectrum of each target species and comparing it to the pooled ``average bird'' (Fig.~\ref{fig:birds-spectrum}), the per-species separation below 1--2\,kHz is negligible and almost all of the discriminating energy sits between 2\,kHz and 8\,kHz, so halving the bin count and dropping the bottom 2.5\,kHz costs almost no useful signal. The shorter window and larger hop then cuts the frame count and the FFT size; together with the bin reduction this leaves an input nearly $4\times$ smaller, a real saving on MCU memory and on the front-end cost (Section~\ref{sec:impl:mel}). All log-mels use a $\log\max(x, 10^{-6})$ compression. The time-domain baselines skip this step and read the 16\,kHz waveform directly, following end-to-end audio models that classify from raw waveforms and so avoid the spectrogram entirely~\cite{abdoli, sang}.

\section{TinyML Models \& Training}
\label{sec:models}

Every model in this paper has one sigmoid output per target species and is trained with binary cross-entropy. The non-target bucket from Section~\ref{sec:datasets:build} is not an extra output channel: a non-target clip is presented to the loss as the all-zeros target vector, and the network has to push every output below threshold. This keeps the head width at exactly $N$ regardless of how the non-target pool is built, and a single model serves both bird-sound detection (any output above threshold) and species classification (argmax over the active outputs, or many birds detected). The width of every convolutional and dense layer in the architectures below scales with a single multiplier $m$, so the same code defines a one-parameter family of operating points.

\subsection{Mel Frontend}

The first TinyChirp~\cite{tinychirp} architecture used a $184\times 80$ log-mel with a 1024-point Hann window, a 256-sample hop ($16$\,ms at $16$\,kHz) and 80 mel bins spanning $80$--$8000$\,Hz; we keep it as \texttt{tinychirp\_mel}. Our \texttt{light\_mel} front-end shrinks every axis at once -- a 512-point window, a 480-sample hop ($30$\,ms) and 40 mel bins restricted to $2500$--$8000$\,Hz -- yielding a $99\times 40$ spectrogram: about $3.7\times$ fewer values and, on-device, $4$ to $5\times$ cheaper to compute (Section~\ref{sec:eval:preproc}).

\subsection{Baseline}
\label{sec:models:baseline}

The starting point are the three TinyChirp~\cite{tinychirp} architectures that fit the MCU resource envelope. We leave out its two SqueezeNet variants, which the TinyChirp evaluation reported as out-of-memory on the target microcontroller, and re-implement the rest with two changes from the reference code: the final Softmax becomes a per-class sigmoid, and the channel multiplier $m$ is wired through every channel count and dense width.

\textbf{CNN-Mel --} Two $3\times 3$ Conv2D blocks with ReLU and $2\times 2$ MaxPooling on the \texttt{tinychirp\_mel} input, flattened into an FC + ReLU head before the $N$-way sigmoid output. $25\,600$ parameters at $m{=}1$.

\textbf{CNN-Time --} Two 1D Conv + ReLU blocks on the $48\,000$-sample waveform, with $2\times 1$ MaxPooling and SpatialDropout between them, an AveragePooling that collapses the time axis to one, and an FC + ReLU head. 748 parameters at $m{=}1$.

\textbf{Transformer-Time --} One 1D Conv + ReLU + MaxPooling stem, a global-average pool that emits a single $16m$-dimensional token, and one pre-norm single-head transformer block (FFN width $32m$), followed by an FC head. 2306 parameters at $m{=}1$ -- slightly above the figure reported in the TinyChirp paper because we match the reference implementation rather than the table in the paper.

\begin{table}[t]
\centering
\scriptsize
\caption{Baseline architectures. Shapes are reported with the channel multiplier $m$ explicit; $N$ is the target species count.}
\label{tab:models-baseline}
\begin{tabular}{lll}
\toprule
Layer & Input Shape & Output Shape \\
\midrule
\multicolumn{3}{l}{\textbf{CNN-Mel} ($25\,600 \cdot m$ params)} \\
\midrule
$3\times 3$ Conv2D + ReLU & $184\times 80\times 1$    & $182\times 78\times 4m$ \\
MaxPooling                & $182\times 78\times 4m$   & $91\times 39\times 4m$ \\
$3\times 3$ Conv2D + ReLU & $91\times 39\times 4m$    & $89\times 37\times 4m$ \\
MaxPooling                & $89\times 37\times 4m$    & $44\times 18\times 4m$ \\
Flatten                   & $44\times 18\times 4m$    & $3168m\times 1$ \\
FC + ReLU                 & $3168m\times 1$           & $8m\times 1$ \\
FC + Sigmoid              & $8m\times 1$              & $N\times 1$ \\
\midrule
\multicolumn{3}{l}{\textbf{CNN-Time} ($748 \cdot m$ params)} \\
\midrule
$3\times 1$ Conv1D + ReLU & $1\times 48\,000$         & $4m\times 48\,000$ \\
MaxPooling                & $4m\times 48\,000$        & $4m\times 24\,000$ \\
SpatialDropout 0.1        & $4m\times 24\,000$        & $4m\times 24\,000$ \\
$3\times 1$ Conv1D + ReLU & $4m\times 24\,000$        & $8m\times 24\,000$ \\
MaxPooling                & $8m\times 24\,000$        & $8m\times 12\,000$ \\
SpatialDropout 0.1        & $8m\times 12\,000$        & $8m\times 12\,000$ \\
Average Pooling           & $8m\times 12\,000$        & $8m\times 1$ \\
FC + ReLU                 & $8m\times 1$              & $64m\times 1$ \\
FC + Sigmoid              & $64m\times 1$             & $N\times 1$ \\
\midrule
\multicolumn{3}{l}{\textbf{Transformer-Time} ($2306 \cdot m$ params)} \\
\midrule
Conv1D + ReLU             & $1\times 48\,000$         & $16m\times 48\,000$ \\
MaxPooling                & $16m\times 48\,000$       & $16m\times 24\,000$ \\
Dropout 0.25              & $16m\times 24\,000$       & $16m\times 24\,000$ \\
Global Average Pooling    & $16m\times 24\,000$       & $16m\times 1$ \\
SingleHeadTransformer     & $16m\times 1$             & $16m\times 1$ \\
FC + Sigmoid              & $16m\times 1$             & $N\times 1$ \\
\bottomrule
\end{tabular}
\end{table}

\subsection{PolyChirp Birdsong Classification Models}
\label{sec:models:classif}

\textbf{Mel-PolyChirp} uses \texttt{light\_mel}; \textbf{DS-CNN} reads \texttt{tinychirp\_mel} with depthwise-separable blocks; \textbf{WrenNet} reads the same spectrogram with markedly richer operators; and \textbf{SincNet} and \textbf{LEAF} drop the mel frontend and learn it directly from the waveform. All five use the same width multiplier $m$ as the baselines. DS-CNN and WrenNet are run with \texttt{tinychirp\_mel} only; pairing every model with both front-ends was out of scope, an arbitrary cap to bound training time and the storage taken by results.

\textbf{Mel-PolyChirp --} Three $3\times 3$ Conv2D + ReLU blocks ($20m$ channels) with a global average pool and a $32m$-wide dense head, run on \texttt{light\_mel}. The goal is to have a model as fast as possible, especially in the case of having the NPU.

\textbf{DS-CNN~\cite{mlperftiny} --} The MLPerf Tiny keyword-spotting reference (Zhang et al.): a stride-2 $10\times 4$ Conv2D stem ($8$ channels) and four depthwise-separable blocks ($3\times 3$ depthwise + $1\times 1$ pointwise, batch-norm and ReLU throughout), a global average pool, and the $N$-way sigmoid. It reads \texttt{tinychirp\_mel}. Unlike the channel-only architectures it scales compound: depth grows with $m$ and width with $\sqrt m$, so parameters still track $m^2$.

\textbf{WrenNet~\cite{wrennet} --} A recent model for multi-species bird classification on low-power loggers. It reads the \texttt{tinychirp\_mel} with the mel bins as channels and is built from PhiNet-style blocks: each block applies a pointwise convolution, a dilated causal depthwise convolution, a squeeze-and-excitation gate, and a pointwise projection, with batch-norm and hard-swish throughout. An initial stage feeds three such blocks and two closing stages, residual connections link the blocks, and a GRU plus a temporal-attention head pool the time axis before the sigmoid output. We keep it to test the limits of the edge path rather than for its footprint. The GRU does not convert to int8 TFLite, where it becomes a custom recurrent op, so the model cannot be baked for microflow or the Axon NPU, and the dilated depthwise convolutions and SE gates lie outside the NPU's supported CNN operators (Section~\ref{sec:impl:npu}). We run it with the GRU replaced by the attention pool, the rest being INT8-compatible.

\textbf{SincNet~\cite{sincnet} --} Forty trainable band-pass sinc filters (mel-initialised over $2$--$8$\,kHz) replace the mel front-end, feeding two strided Conv2D blocks ($24m$ channels) and a flatten + $16m$-dense head; a pairwise band-overlap penalty keeps the filters from collapsing during training. The sinc front-end is hard to quantize, so we add BatchNormalization after every convolution and use ReLU6 in place of ReLU, keeping the per-tensor dynamic range tight enough for the int8 PTQ calibration~\cite{jacob2018quantization} of Section~\ref{sec:impl:ptq}.

\textbf{LEAF~\cite{leaf} --} Learnable Gabor filterbank, Gaussian pooling, and PCEN~\cite{pcen} compression on the raw waveform, feeding two Conv1D + BN + ReLU + MaxPool blocks ($16m$ channels), a global average pool, and a $16m$-wide dense head.

Table~\ref{tab:models-classif} lists the layer-by-layer shapes with $m$ explicit; WrenNet and DS-CNN are omitted, as WrenNet's repeated multi-branch blocks and DS-CNN's compound depth scaling do not fit the flat per-layer format.

\begin{table}[t]
\centering
\scriptsize
\caption{PolyChirp classification architectures. $m$: channel multiplier; $N$: target species count.}
\label{tab:models-classif}
\begin{tabular}{lll}
\toprule
Layer & Input Shape & Output Shape \\
\midrule
\multicolumn{3}{l}{\textbf{Mel-PolyChirp}} \\
\midrule
$3\times 3$ Conv2D + ReLU & $99\times 40\times 1$     & $97\times 38\times 20m$ \\
Average Pooling           & $97\times 38\times 20m$   & $48\times 19\times 20m$ \\
$3\times 3$ Conv2D + ReLU & $48\times 19\times 20m$   & $46\times 17\times 20m$ \\
MaxPooling                & $46\times 17\times 20m$   & $23\times 8\times 20m$ \\
$3\times 3$ Conv2D + ReLU & $23\times 8\times 20m$    & $21\times 6\times 20m$ \\
Global Average Pooling    & $21\times 6\times 20m$    & $20m\times 1$ \\
FC + ReLU                 & $20m\times 1$             & $32m\times 1$ \\
FC + Sigmoid              & $32m\times 1$             & $N\times 1$ \\
\midrule
\multicolumn{3}{l}{\textbf{SincNet}} \\
\midrule
Sinc Conv (40 filt.)            & $1\times 48\,000$   & $40\times 2996$ \\
BN + ReLU6                      & $40\times 2996$     & $40\times 2996$ \\
Average Pooling                 & $40\times 2996$     & $40\times 749$ \\
$8\times 1$ Conv2D + BN + ReLU6 & $40\times 749$      & $24m\times 375$ \\
Average Pooling                 & $24m\times 375$     & $24m\times 93$ \\
$8\times 1$ Conv2D + BN + ReLU6 & $24m\times 93$      & $24m\times 47$ \\
Flatten                         & $24m\times 47$      & $1128m\times 1$ \\
FC + BN + ReLU6                 & $1128m\times 1$     & $16m\times 1$ \\
FC + Sigmoid                    & $16m\times 1$       & $N\times 1$ \\
\midrule
\multicolumn{3}{l}{\textbf{LEAF}} \\
\midrule
Gabor Conv1D (40 filt.)   & $1\times 48\,000$         & $40\times 3000$ \\
Gaussian Pool             & $40\times 3000$           & $40\times 750$ \\
PCEN + BN                 & $40\times 750$            & $40\times 750$ \\
Conv1D + BN + ReLU        & $40\times 750$            & $16m\times 750$ \\
MaxPooling                & $16m\times 750$           & $16m\times 375$ \\
Conv1D + BN + ReLU        & $16m\times 375$           & $16m\times 375$ \\
MaxPooling                & $16m\times 375$           & $16m\times 187$ \\
Global Average Pooling    & $16m\times 187$           & $16m\times 1$ \\
FC + BN + ReLU            & $16m\times 1$             & $16m\times 1$ \\
FC + Sigmoid              & $16m\times 1$             & $N\times 1$ \\
\bottomrule
\end{tabular}
\end{table}

\subsection{Training with Augmentation}
\label{sec:models:train}

All models are trained with Adam and binary cross-entropy on the 70/15/15 split described in Section~\ref{sec:datasets:build}, sampled at natural class proportions. Sample weights are set so that the non-target bucket carries half of the total weight mass and the other half is split between the target classes at inverse frequency; pure inverse-frequency collapses the non-target weight when that bucket outnumbers any single target, which in real life is most of the recorded time.

Audio augmentation is applied to the training stream only and runs on the raw waveform before any mel computation, so every front-end sees the same perturbations:
\begin{itemize}
    \item \textbf{Shift} -- circular shift, up to $\pm 10$\,\% of clip length, $p = 0.5$.
    \item \textbf{Gain} -- uniform gain in $\pm 20$\,dB, $p = 0.5$.
    \item \textbf{Polarity inversion} -- sign flip, $p = 0.5$.
    \item \textbf{Noise} -- with $p = 0.5$, one or two of \{Gaussian noise at amplitude $10^{-4}$ to $3\times 10^{-3}$; AudioSet background mixed at $8$--$25$\,dB SNR\}.
    \item \textbf{Time masking} -- zero a single $2$--$8$\,\% band of the clip, $p = 0.25$.
    \item \textbf{Clipping distortion} -- soft clip between the 0th and 5th amplitude percentiles, $p = 0.10$.
\end{itemize}

\section{TinyML Deployment Pipeline}
\label{sec:pipelines}

Sections~\ref{sec:datasets} and~\ref{sec:models} contribute two reusable parts: a builder that turns a deployment site into a labelled multi-class dataset, and a family of models that read either a log-mel spectrogram or the raw 16\,kHz waveform. What a sensor runs in the field is a particular composition of the two, and the same parts compose in more than one way. This section fixes the two single-stage pipelines we deploy and states the questions the evaluation (Section~\ref{sec:eval}) answers about them.

\label{sec:pipelines:classif}
\textbf{Classification Pipeline --} The default pipeline is site-driven and species-aware. A campaign is specified once, off-device, by a coordinate, a radius and a target count $N$ -- the species of interest. The builder of Section~\ref{sec:datasets:build} pulls and labels that site's audio into $N$ target classes and one pooled non-target bucket, a model from Section~\ref{sec:models} is trained on it, and the quantized network is flashed to the sensor. In the field the network slides over the stream and emits $N$ sigmoid scores per 3\,s window. Because the non-target pool was trained as the all-zeros target (Section~\ref{sec:models}), those scores answer both questions a campaign asks at once: the window is worth storing when any score clears the threshold, and the highest active output names the species. Detection and classification are the same forward pass.

\label{sec:pipelines:detect}
\textbf{Detection Pipeline --} Some campaigns only need to know that a bird was heard, not which one (leaving that to biologists); presence alone decides whether a window is worth keeping. For these the labelling collapses to bird-versus-no-bird and reuses the very same site-specific dataset: every bird clip becomes a positive -- the $N$ target species and the pooled non-target birds alike -- and only the non-bird material, the AudioSet noise and the no-bird windows, stays negative. Within that positive class a few abundant species would otherwise dominate, so before pooling we cap every species at the same maximum number of clips -- up to 250 from each source, 500 in all -- which balances the bird set across species rather than letting the most-recorded birds swamp it. The result is a one-output bird-presence detector, the multi-species counterpart of the single-species screen of TinyChirp~\cite{tinychirp}, and the cheapest model we deploy: a smaller head, a single threshold, no per-species calibration. This is a strictly broader question than the classification pipeline answers -- a non-target bird counts as a positive for the detector, yet maps to the all-zeros target for the classifier -- so the two are distinct models, not two readings of one. The detector stands alone when species identity is not needed.

\section{Implementation Overview}
\label{sec:impl} 

We target two off-the-shelf low-power boards representative of hardware with/without NPU acceleration: respectively the Nordic nRF54LM20 and the Raspberry Pi Pico 2. Both run the models on their Cortex-M core. The nRF54LM20 additionally provides an Axon NPU. 
The same trained network thus alternatively uses two categories of execution paths: portable CPU paths that runs on either board, or an accelerated NPU path (only on the nRF54LM20).

\textbf{CPU Paths --} On all the hardware, we use a Rust firmware built on Ariel OS~\cite{arielos}. For transpiling models, we use two approaches. On the one hand, inference handled by microflow-rs~\cite{microflow}, which turns a quantized TFLite file into a model at build time. The network is generated directly from the \texttt{.tflite}, so the width-multiplier family of Section~\ref{sec:models} compiles into firmware without any hand-written kernels, and the mel front-end (FFT and filterbank) runs in the same Rust path. On the other hand, for comparison, we also implement an alternative CPU path based on Ariel-ML and the IREE transpiler~\cite{iree}. We chose IREE as comparison point as prior work~\cite{arielml} measured improvements compared to TinyChirp~\cite{tinychirp}. Thereafter, such CPU paths are simply marked as ``IREE".

\label{sec:impl:npu}
\textbf{NPU Path --} On the nRF54LM20, we can leverage the Axon NPU 
using its Edge AI add-on~\cite{nrfedgeai}. The add-on's compiler turns the int8 model (Section~\ref{sec:impl:ptq}) into a C header that the Axon driver executes, while the mel spectrogram is still computed on the Cortex-M core and only the resulting feature map is handed to the NPU. 
Note that not every model can take this path. The Axon compiler caps each tensor dimension at 1024 and restricts convolutions to small kernels (filter width up to 32, stride up to 31), and its operator set covers standard CNN blocks -- convolution, pooling, fully connected, and the ReLU family -- but none of the learnable audio front-ends. The mel-spectrogram models fit this envelope naturally: the spectrogram is a compact image ($184\times 80$ for \texttt{tinychirp\_mel} models, $99\times 40$ for \texttt{light\_mel} used in Mel-PolyChirp), well inside the size limit, and its convolutional blocks map directly onto the NPU. The time-domain models do not. A $48\,000$-sample waveform already exceeds the 1024-long input limit, and the band-pass filters of SincNet and the Gabor/PCEN front-end of LEAF are long kernels with no counterpart in the supported operators. These models therefore stay on the CPU. In effect the NPU accelerates the spectrogram branch of the pipeline, while the waveform branch -- which exists precisely to skip the mel computation -- remains a CPU-only option.

\label{sec:impl:ptq}
\textbf{Optimization Techniques --} 
After training in float, we apply post-training quantization (PTQ)~\cite{jacob2018quantization, nagel2021white}, mapping weights and activations to int8 with per-tensor scales and zero-points calibrated on a sample of the training set. 
The Axon NPU executes int8 only, and on the Cortex-M core quantizing to int8 arithmetic keeps weights and activation arena inside the RAM budget. 
As the network infers on a $3\,$s sliding window over the audio stream, consecutive windows overlap. Exploiting this fact, we can use partial convolution as described in TinyDéjàVu~\cite{tinydejavu} to limit peak RAM usage. 
Note that the time-domain models cannot be executed on the NPU  (Section~\ref{sec:impl:npu}), hence they only have a CPU path. 

\label{sec:impl:mel}
\textbf{Mel Front-end --} The mel-spectrogram models rest on a log-mel front-end that, on a microcontroller, is a bottleneck in its own right: Nordby measured mel feature extraction at $60\,$ms against the $38$--$81\,$ms taken by the networks it feeds~\cite{nordby2019esc}, and TinyChirp cites this very cost when it argues for skipping the spectrogram and reading the waveform directly~\cite{tinychirp}. We keep the front-end on-device but make it cheap enough not to dominate the inference budget, by precomputing everything that does not depend on the incoming audio and reusing it across every frame and every sliding window. The Hann window is tabulated once at start-up instead of being re-evaluated for each frame, and the triangular mel filterbank is built once from its band edges and stored sparsely, as a flat list of (FFT bin, mel bin, weight) triples that records only the non-zero filter overlaps. Because each FFT bin falls under at most two adjacent triangles, applying the filterbank then becomes a single pass of about two multiply-accumulates per bin, rather than the dense matrix product across all FFT bins and mel channels that a literal filterbank would run. The FFT is the one step we cannot precompute (a 512-point transform for Mel-PolyChirp, 1024 for the heavier CNN-Mel): on the nRF54LM20 the 512-point transform is offloaded to the Axon NPU, leaving the Cortex-M core only the windowing, magnitude, filterbank, and log; the 1024-point transform exceeds the NPU limit and falls back to a CMSIS FFT, as on every board without the accelerator.

\section{Experimental Performance Evaluation}
\label{sec:eval}

\label{sec:pipelines:rq}

Across all pipelines and models, our evaluation is organised to tackle four axes defined as follows.
\begin{itemize}
    \item \textbf{Accuracy --} Can a single multi-class network both detect a bird and identify which of up to $N=10$ target species it is, and how much of that accuracy survives the int8 post-training quantization the hardware forces (Section~\ref{sec:impl:ptq})?
    \item \textbf{Front-end --} Does reading the raw waveform, and so skipping the mel computation entirely, trade accuracy for a cheaper pipeline against the mel-spectrogram front-end?
    \item \textbf{Budget and hardware --} Does the multi-class setup fit the flash, RAM, latency and energy envelope of a low-power sensor, and what does the on-chip NPU buy across the two boards?
    \item \textbf{Pipeline shape --} When only presence matters, does the species-agnostic detector lower the average on-device cost against always running the full classifier?
\end{itemize}

We report below on measurements we performed on two low-power boards: the Nordic nRF54LM20 and the Raspberry Pi Pico 2. Note that on-device timings are rounded to the nearest millisecond, since the variation between runs is already larger than that.

\subsection{Front-end Preprocessing Cost}
\label{sec:eval:preproc}

We time the mel front-end per 3\,s window on both boards, by step, averaged over ten windows (Table~\ref{tab:preproc}).

Precomputing everything but the FFT (Section~\ref{sec:impl:mel}) makes the front-end $7$ to $11\times$ faster than TinyChirp's literal per-frame version~\cite{tinychirp}; the gain is in framing, where tabulating the Hann window once removes a per-sample cosine that is costly on an MCU. \texttt{light\_mel} is a further $4$ to $5\times$ cheaper than \texttt{tinychirp\_mel}, the FFT and magnitude dominating throughout. The Axon NPU runs the $512$-point FFT of \texttt{light\_mel} about $1.8\times$ faster than the Cortex-M ($30$ vs $54$\,ms), cutting the whole front-end from $126$ to $53$\,ms; it cannot take the $1024$-point transform of \texttt{tinychirp\_mel} (Section~\ref{sec:impl:npu}), which therefore stays on the CPU. That the front-end is a real cost, not just inference, matches earlier keyword-spotting work, which reports roughly one second to compute a much smaller ($30\times40$) spectrogram on a more powerful STM32F7~\cite{kws}.

\begin{table}[t]
\centering
\scriptsize
\setlength{\tabcolsep}{3.5pt}
\caption{Mel front-end cost per 3\,s window, by step (ms);  Rows with \emph{+NPU}: FFT offloaded to Axon NPU. Column \emph{Energy}: total per-window front-end cost (mJ), measured on the nRF54.}

\label{tab:preproc}
\csvreader[respect underscore, tabular=llrrrrrrr,
  table head=\toprule Front-end & Target & Frame & FFT & Mag & Mel & Log & Total & Energy \\\midrule,
  table foot=\bottomrule]%
  {numeric_data/preprocess.csv}%
  {frontend=\fe,target=\tg,frame=\fr,fft=\ff,mag=\mg,mel=\ml,log=\lg,total=\tot,energy=\en}%
  {\texttt{\fe} & \tg & \fr & \ff & \mg & \ml & \lg & \tot & \en}
\end{table}

\subsection{Predictive Performance}
\label{sec:eval:metrics}

Per-class thresholds are tuned on train+val, separately for $F_1$ and $F_2$, and reported on the held-out test split; classification metrics are macro-averaged over the target species, detection metrics are read on the single bird-presence class. Everything is INT8 unless stated, at the $F_2$ operating point, where recall is favoured over precision.

\subsubsection{Classification pipeline}
\begin{figure*}[t]
\centering
\includegraphics[width=\linewidth]{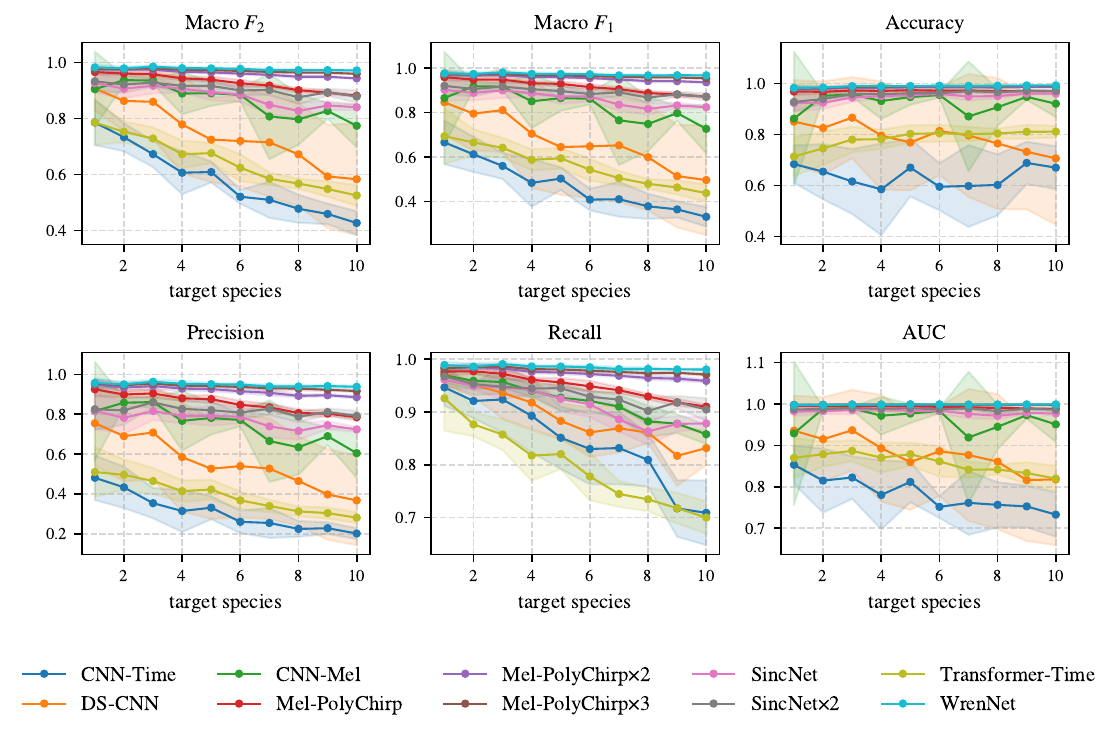}
\caption{Classification predictive performance vs. number of target species $k$ (INT8). Precision and recall are read at the $F_2$-tuned threshold; each curve is a mean over resampled species draws. \texttt{LEAF} collapses under INT8 (Fig.~\ref{fig:lossdrop}) and is excluded.}
\label{fig:scaling}
\end{figure*}

The site-driven classifier grows with the number of target species $k$, so we report it as a scaling curve. Figure~\ref{fig:scaling} tracks the six TinyChirp~\cite{tinychirp} metrics -- macro $F_2$, $F_1$, accuracy, precision, recall, AUC -- from $k{=}1$ to $k{=}10$, each point a mean over resampled species draws.

Quality falls off gradually with $k$ and the models spread apart. The spectrogram models hold up best -- WrenNet\footnote{Throughout, WrenNet~\cite{wrennet} denotes our INT8-compatible variant with the GRU replaced by an attention pool (Section~\ref{sec:models:classif}).} drops only from $0.98$ to $0.97$ macro $F_2$, Mel-PolyChirp×2 stays above $0.94$ -- while the time-domain baselines fare worst, CNN-Time and Transformer-Time halving to $0.43$ and $0.52$. AUC and accuracy stay nearly flat throughout, so the loss is a threshold effect, not lost separability: at the $F_2$ point precision falls first while recall is held high. The run-to-run spread splits the models the same way: Mel-PolyChirp×3 and Mel-PolyChirp×2 hold their macro $F_2$ standard deviation between $0.001$ and $0.006$ across $k{=}1$ to $10$, while DS-CNN and CNN-Mel vary from $0.013$ to $0.200$, over $8$ resampled species draws per point.

\subsubsection{Detection pipeline}
\begin{figure}[t]
\centering
\includegraphics[width=\linewidth]{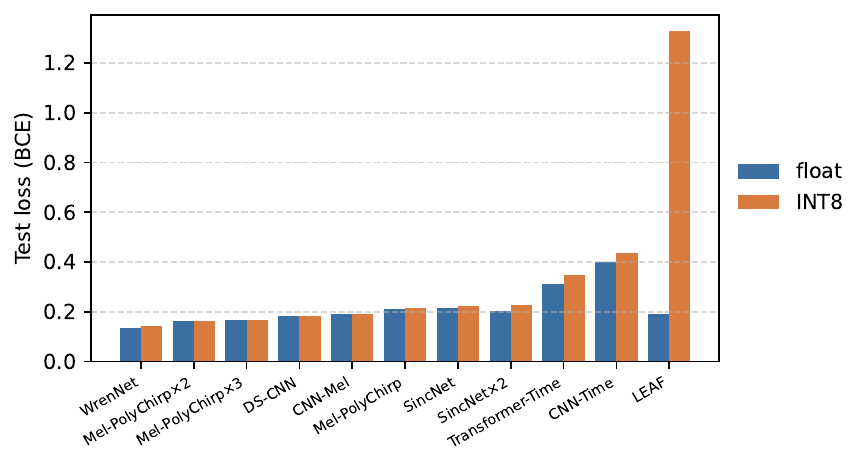}
\caption{Float vs. INT8 test loss per model on detection.}
\label{fig:lossdrop}
\end{figure}

Bird-presence detection is a single binary task, not a scaling one, so we report it once per model (Table~\ref{tab:detection}, INT8 at the $F_2$ point). Every model clears $F_2 \geq 0.91$ at recall $\geq 0.97$: almost no bird window is dropped. WrenNet, the Mel-PolyChirp family and DS-CNN lead, and even the $1$--$3$\,KB time-domain baselines reach $F_2 = 0.91$--$0.92$. LEAF is excluded (Fig.~\ref{fig:lossdrop}).

\begin{table}[t]
\centering
\scriptsize
\caption{Detection predictive performance (INT8, $F_2$-tuned threshold).}
\label{tab:detection}
\begin{filecontents*}{numeric_data/detection.csv}
model,flash_kb,accuracy,precision,recall,f1,f2,auc
CNN-Time,1.17,0.79,0.72,0.97,0.83,0.91,0.90
Transformer-Time,2.65,0.81,0.74,0.97,0.84,0.92,0.93
CNN-Mel,25.1,0.91,0.85,1.00,0.92,0.96,0.97
Mel-PolyChirp,8.39,0.91,0.86,0.99,0.92,0.96,0.97
Mel-PolyChirp×2,32,0.92,0.88,0.99,0.93,0.97,0.98
Mel-PolyChirp×3,70.8,0.93,0.89,0.99,0.94,0.96,0.98
SincNet,32.8,0.88,0.81,1.00,0.90,0.95,0.97
SincNet×2,107,0.88,0.82,0.99,0.90,0.95,0.97
WrenNet,28.2,0.93,0.89,0.99,0.94,0.97,0.99
DS-CNN,1.27,0.92,0.87,0.99,0.93,0.96,0.97
\end{filecontents*}
\csvreader[
  respect underscore,
  tabular=lrrrrrr,
  table head=\toprule Model & Acc. & Prec. & Rec. & $F_1$ & $F_2$ & AUC \\\midrule,
  table foot=\bottomrule
]{numeric_data/detection.csv}{model=\m,accuracy=\ac,precision=\pr,recall=\rc,f1=\fa,f2=\fb,auc=\au}%
{\texttt{\m} & \ac & \pr & \rc & \fa & \fb & \au}
\end{table}

Quantization to INT8 matters for every model but especially for the time-domain ones -- LEAF, CNN-Time, Transformer-Time, and even SincNet, though the latter is designed to resist it (Fig.~\ref{fig:lossdrop}): float and INT8 test loss differ by under $0.4$ everywhere except LEAF, whose loss jumps from $0.19$ to $1.33$ as the learned PCEN gains do not survive a single activation scale, so it is excluded from the quantized comparisons.

\textbf{Label provenance and limitations --} These scores measure agreement with BirdNET, not with ground truth: a clip becomes a target only when BirdNET is at least $0.92$ confident, an arbitrary threshold from prior work~\cite{tinychirp}, so the models learn to copy BirdNET -- labelling mistakes included -- and the test split, drawn from the same recordings, shares that bias. The high threshold keeps the labels clean but narrows the data to the loud, clear calls, not the faint or overlapping ones that fill a real recording. Distilling from BirdNET's confidence scores, as WrenNet does~\cite{wrennet}, would keep some of that information and reuse the audio we currently drop. The evaluation is also confined to one site (Paris, $50\,$km) and to curated Xeno-Canto and Macaulay clips rather than audio captured by the target sensor in the field; some target species are under-represented for lack of data, so the rarer classes train on less. More sources, other sites, and on-device field recordings would close the gap between this curated set and what the device hears.

\subsection{Inference Cost and Latency}
\label{sec:eval:inference}

We deploy every model on the nRF54LM20 and report its inference cost there (Table~\ref{tab:nrf-inference}) under each runtime it can target: the \texttt{microflow} INT8 transpiler, with and without its streaming kernel functionality; the IREE ahead-of-time compiler; and the on-chip neural accelerator (NPU). To place those numbers in context we time a handful of models on the RP2350 (Pico~2) as well, giving the cross-platform factors we use to reason about the rest (Table~\ref{tab:calibration}).

\begin{table}[t]
\centering
\footnotesize
\caption{Cross-platform latency calibration, 
Clocks: Pico~2 $150$\,MHz, nRF54 $128$\,MHz. MF~=~\texttt{microflow}, MF-S~=~its streaming kernel (used where the non-streaming path is OOM on a board).}
\label{tab:calibration}
\csvreader[respect underscore, tabular=llrr,
  table head=\toprule & & \multicolumn{2}{c}{Latency (ms)}\\\cmidrule(lr){3-4} Model & Mode & Pico2 & nRF54\\\midrule,
  table foot=\bottomrule]%
  {numeric_data/calibration.csv}%
  {model=\m,mode=\mo,lat_mf_pico=\lp,lat_mf_nrf=\ln}%
  {\texttt{\m} & \mo & \lp & \ln}
\end{table}

On the models that fit both runtimes, \texttt{microflow} runs about $2.5\times$ faster on the Pico~2 ($150$\,MHz) than on the nRF54 ($128$\,MHz); the $1.17\times$ clock ratio explains only part of this, the rest being a consistent platform factor -- memory-access latency, as both boards carry $512$\,KiB of RAM -- that we fold into the calibration.

\begin{figure}[t]
\centering
\includegraphics[width=\linewidth]{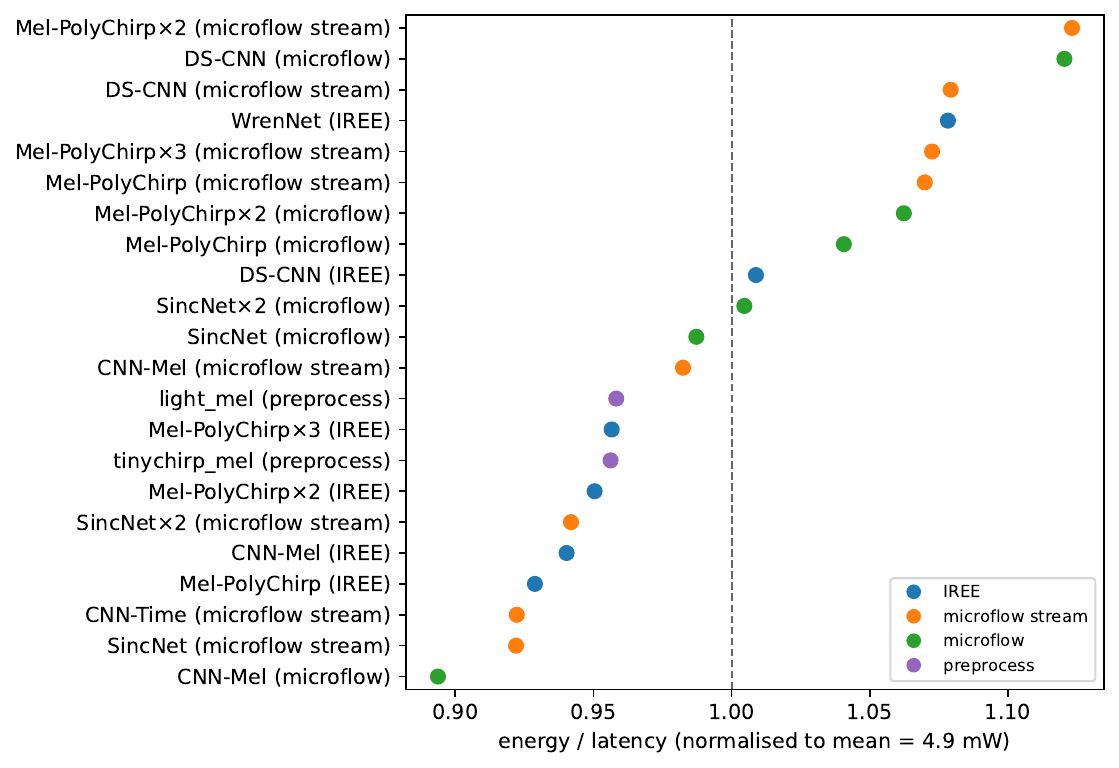}
\par\smallskip
\includegraphics[width=\linewidth]{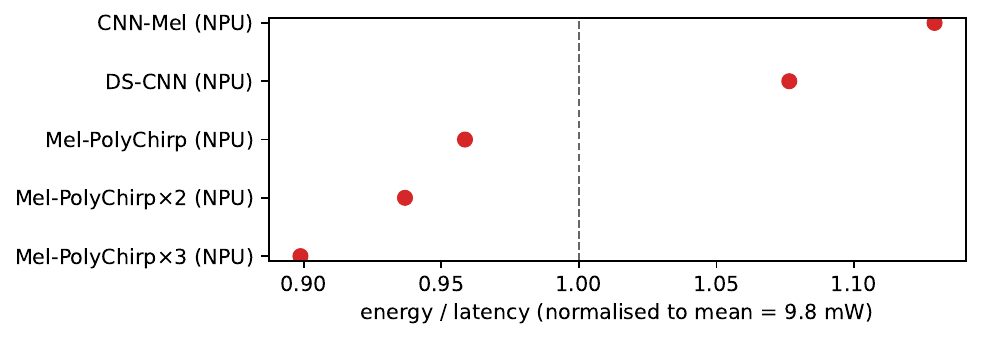}
\caption{Effective power (energy over latency) measured on nRF54LM20. Top: on MCU. Bottom: on NPU. Normalised to group mean (dashed line). The three CPU runtimes cluster near $4.9$\,mW, the NPU near $9.8$\,mW -- so energy per inference is essentially latency times a per-regime constant, and the two regimes differ mainly in latency.
}
\label{fig:power-ratio}
\end{figure}

On the nRF54 itself (Table~\ref{tab:nrf-inference}) the NPU dominates -- CNN-Mel in $2$\,ms and Mel-PolyChirp×3 in $39$\,ms -- whereas the transpiler runs from hundreds of milliseconds up to several seconds, and its non-streaming path runs out of memory at the widest width (``OOM'') where streaming, which never materialises the full feature maps, still fits. Note that time-domain models have neither a compiled nor an NPU path (IREE provides no streaming kernel, so the raw $48\,000$-sample input overflows memory, while the NPU does not take a raw waveform). The LEAF and Transformer-Time models could not be run on any backend: the PCEN mechanism is not runnable on int8 models, while for Transformer-Time the first layer ($16\times 48\,000$) cannot exist without streaming, which microflow does not provide for the attention mechanism. The only time models measured (\texttt{CNN-Time}, \texttt{SincNet}) therefore run under \texttt{microflow} alone -- \texttt{SincNet} in both modes, while \texttt{CNN-Time} fits only with streaming. Energy per inference (right half of each table) closely tracks latency, along two lines rather than one (Fig.~\ref{fig:power-ratio}): the three runtimes draw comparable power (4.9 mW), so their energy is roughly latency times a shared constant, whereas the NPU sits in a separate regime, drawing about twice that power (9.8 mW) but finishing inference one to two orders of magnitude faster, thus spending far less energy per inference. Taking the $3\,$s window as the real-time bound (a real-time factor of latency$/3000$ below $1$ is needed to keep pace with the stream), the NPU clears it for every mel model by two orders of magnitude, and the smaller CPU configurations (CNN-Mel, Mel-PolyChirp, DS-CNN, single-width SincNet and CNN-Time) also stay under $1$; but the ×2/×3 mel widths and SincNet×2 do not -- Mel-PolyChirp×3 reaches $4.9\,$s streaming and $5.5\,$s under IREE, already before the $126\,$ms front-end -- so on a single Cortex-M core these cannot sustain a continuous stream and require the NPU or window skipping.\ \\
\indent A four-point spot-check on the Pico~2, spanning $74$\,ms to $1.6$\,s across inference and preprocessing, finds the effective power (energy over latency) constant at $70$\,mW to within $14\%$, so its board-level energy follows a single $E = P \times \text{latency}$ law and needs no per-model table. Measured on the full $3.3$\,V rail, for the recorded operations the Pico takes $1.3$ to $2.2\times$ less time than the nRF54 but draws $6.5$ to $12\times$ more energy, making it suitable only where energy is not constrained.

Flash follows the same runtime ordering (Table~\ref{tab:nrf-inference}, right): the \texttt{microflow} transpiler yields the smallest binaries, while IREE's ahead-of-time kernels and runtime add roughly $100$\,kB. All three land above comparable RIOT-OS deployments such as TinyChirp~\cite{tinychirp}, as the footprint is set less by the model than by the Rust-based Ariel~OS base firmware, which is heavier than RIOT's.

\newcommand{\nrfhead}{\toprule & \multicolumn{4}{c}{Inference Latency (ms)} & \multicolumn{4}{c}{Energy (mJ)} & \multicolumn{4}{c}{Flash (kB)}\\\cmidrule(lr){2-5}\cmidrule(lr){6-9}\cmidrule(lr){10-13} Model & IREE & microflow & \makecell{microflow\\w/ streaming} & NPU & IREE & microflow & \makecell{microflow\\w/ streaming} & NPU & IREE & microflow & \makecell{microflow\\w/ streaming} & NPU\\\midrule}
\newcommand{\nrftimehead}{\toprule & \multicolumn{2}{c}{Inference Latency (ms)} & \multicolumn{2}{c}{Energy (mJ)} & \multicolumn{2}{c}{Flash (kB)}\\\cmidrule(lr){2-3}\cmidrule(lr){4-5}\cmidrule(lr){6-7} Model & microflow & \makecell{microflow\\w/ streaming} & microflow & \makecell{microflow\\w/ streaming} & microflow & \makecell{microflow\\w/ streaming}\\\midrule}

\begin{table*}[t]
\centering
\footnotesize
\setlength{\tabcolsep}{3.5pt}
\caption{On-device inference cost on the nRF54LM20. All models quantized to INT8 precision. Time-domain models run under \texttt{microflow} only; \texttt{CNN-Time} fits only with streaming. Flash is the binary size of the deployable firmware. OOM: model can fit flash but exceeds available RAM at run time; NA: model not supported by runtime due to insufficient operator coverage.}
\label{tab:nrf-inference}
\emph{Mel-spectrogram models}\par\smallskip
\csvreader[respect underscore, tabular=lrrrrrrrrrrrr, table head=\nrfhead, table foot=\bottomrule]%
  {numeric_data/nrf_mel.csv}%
  {model=\m,lat_iree=\ai,lat_mfs=\as,lat_mf=\af,lat_npu=\an,e_iree=\ei,e_mfs=\es,e_mf=\ef,e_npu=\eno,flash_iree=\fli,flash_mf=\flf,flash_mfs=\fls,flash_npu=\fln}%
  {\texttt{\m} & \ai & \af & \as & \an & \ei & \ef & \es & \eno & \fli & \flf & \fls & \fln}
\par\medskip
\emph{Time-domain models} (\texttt{microflow})\par\smallskip
\csvreader[respect underscore, tabular=lrrrrrr, table head=\nrftimehead, table foot=\bottomrule]%
  {numeric_data/nrf_time.csv}%
  {model=\m,lat_mf=\af,lat_mfs=\as,e_mf=\ef,e_mfs=\es,flash_mf=\flf,flash_mfs=\fls}%
  {\texttt{\m} & \af & \as & \ef & \es & \flf & \fls}
\end{table*}

\newcommand{\picohead}{\toprule Model & IREE & microflow & \makecell{microflow\\w/ streaming}\\\midrule}
\newcommand{\picotimehead}{\toprule Model & microflow & \makecell{microflow\\w/ streaming}\\\midrule}

\begin{table}[t]
\centering
\footnotesize
\caption{On-device inference latency (ms) on the RP2350 (Pico~2). All models quantized to INT8 precision. OOM: model can fit flash but exceeds available RAM at run time; NA: model not supported by runtime due to insufficient operator coverage.}
\label{tab:pico-inference}
\emph{Mel-spectrogram models}\par\smallskip
\csvreader[respect underscore, tabular=lrrr, table head=\picohead, table foot=\bottomrule]%
  {numeric_data/pico_mel.csv}%
  {model=\m,lat_iree=\ai,lat_mfs=\as,lat_mf=\af}%
  {\texttt{\m} & \ai & \af & \as}
\par\medskip
\emph{Time-domain models} (\texttt{microflow})\par\smallskip
\csvreader[respect underscore, tabular=lrr, table head=\picotimehead, table foot=\bottomrule]%
  {numeric_data/pico_time.csv}%
  {model=\m,lat_mf=\af,lat_mfs=\as}%
  {\texttt{\m} & \af & \as}
\end{table}

\section{Discussion \& Perspectives}


\textbf{Hardware Accelerator (NPU) Considerations --} Using an NPU remains somewhat challenging. At the time of writing, this requires using a separate C firmware on the nRF Connect SDK (Zephyr) combined with Nordic's closed-source Edge AI add-on~\cite{nrfedgeai}. The add-on is easy to use, but we experienced downsides. First, this path departs from the single-language (Rust and Ariel OS) stack described in Section~\ref{sec:impl}. Second, it ties the deployment to a specific vendor's hardware. However, the Axon NPU does execute much faster the convolutional part of mel models compared to Cortex-M cores. Inference latency decreases 2 orders of magnitude (from about a second to a few milliseconds, recall Table~\ref{tab:nrf-inference}), and inference energy consumption drops proportionally, which is a strong motivation. We observe furthermore how the front-end benefits from the NPU. The accelerator supports ML layers, but also direct calls to operations like FFT. 
With the network offloaded, the mel computation takes most of the time spent on each window. Waveform models would gain the most from NPU, since they are almost all convolution. However, the NPU cannot run these models, as a $48\,000$-sample input is over the 1024 limit on tensor dimensions, and the Gabor front-end is not supported. Hence, NPU hardware limitations would have to be lifted in next-generation NPUs.

\textbf{Model Transpilers --}
The choice of model transpiler (we studied microflow and IREE) is a system-level engineering decision involving trade-offs among inference latency, memory usage, operator coverage, power efficiency, etc. For instance, compared with microflow, we measured IREE incurs a substantially ($>2 \times$) larger Flash runtime footprint and less memory-safety guarantees, but uses less RAM, achieves less latency and provides support for a broader range of model operators. Orthogonally, given the significant acceleration capability and power efficiency of NPUs we measured, future work should further explore adding NPU support to both IREE and microflow. Future work should also develop further operator support for microflow. Overall, the benchmarks presented in this work may provide useful reference points for practitioners aiming to select a suitable transpiler strategy.


\textbf{Recall \textit{versus} Precision Trade-offs --} 
A false negative drops a window for good and loses a real detection. A false positive only keeps a spurious clip, and we can filter it out later when the SD card is read. Hence we favour recall over precision, and tune per-class thresholds for $F_2$ as well as $F_1$ (see Section~\ref{sec:eval:metrics}). Quantization complicates this. PTQ to int8 shifts the scores, so a threshold set on the float model must be ``retuned" per backend (see Section~\ref{sec:impl:ptq}). There is no single right threshold, since it depends on how fast the SD card fills and how badly a campaign can afford to miss a species. The detection pipeline uses the cheapest case, with one threshold set for recall and the species left to the offline pass. The harder problem comes earlier. We weight the training set so the loss does not collapse onto the large non-target class (Section~\ref{sec:models:train}), but the field distribution is not the one we trained on, so a threshold that looks right on the test set can behave differently in deployment.


\textbf{Batching Inference Under a Larger RAM Budget --} The limiting factor is often RAM, and the streaming scheme of Section~\ref{sec:impl:ptq} is there to keep it low. If a board has more RAM to spare, we could instead run several 3\,s windows through the network at once. Reusing each weight across a batch, while in cache, would lower the per-call overhead and the energy spent per window. This seems like a low-hanging fruit on larger boards where RAM is not the limiting factor (as this costs extra activation memory).

\section{Related Work}

{\bf Birdsong recognition on edge \& server hardware --}
The main reference for automatic bird-sound recognition is BirdNET~\cite{birdnet}, a residual network that identifies several thousand species from continuous soundscape audio.
BirdNET-Pi~\cite{birdnetpi} runs a quantized BirdNET continuously on a Raspberry Pi to log detections in real time. 
More recent work such as Perch~\cite{perch} sharpens the underlying representation with general-purpose birdsong embeddings that transfer well across tasks.
On the other hand, AudioMoth~\cite{audiomoth} autonomous recording units 
records to an SD card, but offloads species recognition to an offline step.
Across this domain, the classification runs only on microprocessor- or GPU-class hardware.

{\bf Birdsong recognition on microcontrollers --}
Early on-device detectors pair a mel front-end with a compact CNN: Disabato et al.~\cite{disabato} detect birdsong at the edge on a Cortex-M, TinyBird-ML~\cite{tinybird} runs vocalization analysis and syllable classification on an animal-borne ultra-low-power node, and Miquel et al.~\cite{miquel} study the energy cost of edge audio for biologging. WrenNet~\cite{wrennet}  distills BirdNET into a small recurrent network. TinyChirp~\cite{tinychirp} uses a Cortex-M to screen a single target species in the field 
while TinyD\'ej\`aVu~\cite{tinydejavu} lowers the RAM of streaming inference for such sensors. 

{\bf TinyML pipelines --}
TensorFlow Lite Micro~\cite{tflitemicro} is the common C++ interpreter, usually paired on Arm Cortex-M with the CMSIS-NN kernel library~\cite{cmsisnn}. 
Transpilers such as $\mu$TVM~\cite{tvm} or IREE~\cite{iree} lower a neural network output from PyTorch or tensorFlow to low-level code for various microcontrollers. MCUNet~\cite{mcunet} combines such an engine with neural architecture search to fit the model and runtime to a given memory budget. 
U-TOE~\cite{utoe}, RIOT-ML~\cite{riotml} or Ariel-ML~\cite{arielml} combine transpilers and embedded OS integration, so that arbitrary models can be flashed, benchmarked, and updated over low-power network links on various boards. 
Another approach is Microflow-rs~\cite{microflow}, which compiles quantized models into memory-safe Rust at build time -- an interesting perspective as embedded Rust is now a reasonable alternative to C for software on microcontrollers~\cite{thapa2026embedded-rust}. 
Vendor pipelines also include STM32Cube.AI and Edge Impulse. 
Orthogonally, techniques are developed to lower peak RAM usage. For instance msf-CNN~\cite{msfcnn} leverages patch-based layer-fusion. 

{\bf Microcontroller hardware \& neural accelerators --}
Popular microcontrollers are based on 32-bit instruction set architectures such as Arm Cortex-M, RISC-V 32 or Xtensa. Alongside such architectures, a class of microcontroller-scale neural accelerators ($\mu$NPUs) has appeared: Arm's Ethos-U55~\cite{ethosu55}, the Maxim MAX78000~\cite{max78000}, ST's STM32N6 with its Neural-ART engine~\cite{stm32n6}, and the Nordic Axon we use on the nRF54LM20~\cite{nrfedgeai}. These NPUs cut inference time and energy by large factors, but run int8 only, cover a restricted set of CNN operators, and so far depend on vendor toolchains, which limits both model coverage and portability across boards. 
Standardized TinyML benchmark suites such as MLPerf Tiny~\cite{mlperftiny} target plain MCU cores, while Millar et al.~\cite{unpu} recently provided cross-vendor benchmarks of several $\mu$NPUs and find measured performance often departs from the data sheets. 

\section{Conclusion}

Bio-acoustic monitoring leveraging TinyML pipelines is an approach that is appealing and which has recently become practical. In this paper, we have designed, implemented and evaluated PolyChirp, a low-power embedded software framework which enables training and baking different TinyML models in the firmware of various microcontroller-based hardware, with or without NPU hardware acceleration. Compared to prior work which either required microprocessor-class resources or performed only single-species detection, PolyChirp can classify with high accuracy up to 10 species simultaneously on common microcontroller hardware. We also measure how available NPU hardware acceleration is best leveraged on such hardware. Combined with the portable open source code implementations we published, PolyChirp provides a solid base for both practical deployments and further research in this field.
\ \\ \ \\
\textbf{Code Availability --}
The pipeline and benchmark code are published at \href{https://github.com/ariel-os/ariel-microflow-ml}{github.com/ariel-os/ariel-microflow-ml} and at \href{https://github.com/NathanDuboisset/polychirp}{github.com/NathanDuboisset/polychirp} respectively.
\ \\ \ \\
\textbf{Acknowledgements --} This work was partly funded by the France 2030 ANR program “PEPR Networks of the Future” (ref. ANR-22-PEFT-0003, ANR-22-PEFT-0007).

\bibliographystyle{IEEEtran} 
\bibliography{main}

@inproceedings{tinychirp,
  author = {Huang, Zhaolan and others},
  title = {{TinyChirp}: Bird Song Recognition Using {TinyML} Models on Low-power Wireless Acoustic Sensors},
  booktitle = {IEEE IS2},
  year = {2024}
}

@inproceedings{tinydejavu,
  author = {Huang, Zhaolan and Baccelli, Emmanuel},
  title = {{TinyD\'ej\`aVu}: Smaller {RAM} and Faster Inference with Neural Networks on {MCUs} for Sensor Data Streams},
  booktitle = {IEEE DCOSS-IoT Workshop (ISIoT)},
  year = {2026},
}

@article{birdnet,
  author = {Kahl, Stefan and others},
  title = {{BirdNET}: A deep learning solution for avian diversity monitoring},
  journal = {Ecological Informatics},
  volume = {61},
  pages = {101236},
  month = mar,
  year = {2021},
  doi = {10.1016/j.ecoinf.2021.101236}
}

@inproceedings{arielos,
  author = {Frank, Elena and others},
  title = {{Ariel OS}: An Embedded {Rust} Operating System for Networked Sensors \& Multi-Core Microcontrollers},
  booktitle = {Proc. IEEE DCOSS-IoT},
  month = jun,
  year = {2025},
  note = {arXiv:2504.19662}
}

@mastersthesis{nordby2019esc,
  author = {Nordby, Jon},
  title = {Environmental Sound Classification on Microcontrollers using Convolutional Neural Networks},
  school = {NMBU},
  year = {2019},
  note = {\url{http://hdl.handle.net/11250/2611624}}
}

@article{microflow,
  author = {Carnelos, Matteo and others},
  title = {{MicroFlow}: An Efficient {Rust}-Based Inference Engine for {TinyML}},
  journal = {Internet of Things},
  volume = {30},
  pages = {101498},
  year = {2025}
}

@misc{nrfedgeai,
  author = {{Nordic Semiconductor}},
  title = {Edge {AI} Add-on for {nRF} Connect {SDK}},
  howpublished = {\url{https://nrfconnectdocs.nordicsemi.com/addons/addon-edge-ai/latest/index.html}},
  note = {Accessed: 2026-08-14}
}

@inproceedings{jacob2018quantization,
  author = {Jacob, Benoit and others},
  title = {Quantization and Training of Neural Networks for Efficient Integer-Arithmetic-Only Inference},
  booktitle = {Proc. IEEE CVPR},
  year = {2018}
}

@article{nagel2021white,
  author = {Nagel, Markus and others},
  title = {A White Paper on Neural Network Quantization},
  journal = {arXiv preprint arXiv:2106.08295},
  year = {2021}
}

@inproceedings{sincnet,
  author = {Ravanelli, Mirco and Bengio, Yoshua},
  title = {Speaker Recognition from Raw Waveform with {SincNet}},
  booktitle = {Proc. IEEE Spoken Language Technology Workshop (SLT)},
  pages = {1021--1028},
  year = {2018}
}

@inproceedings{leaf,
  author = {Zeghidour, Neil and others},
  title = {{LEAF}: A Learnable Frontend for Audio Classification},
  booktitle = {Proc. ICLR},
  year = {2021}
}

@inproceedings{pcen,
  author = {Wang, Yuxuan and others},
  title = {Trainable Frontend for Robust and Far-Field Keyword Spotting},
  booktitle = {Proc. IEEE ICASSP},
  pages = {5670--5674},
  year = {2017}
}

@inproceedings{audioset,
  author = {Gemmeke, Jort F. and others},
  title = {Audio Set: An Ontology and Human-Labeled Dataset for Audio Events},
  booktitle = {IEEE ICASSP},
  year = {2017}
}

@misc{xenocanto,
  author = {{Xeno-canto Foundation}},
  title = {xeno-canto: Sharing Wildlife Sounds from Around the World},
  howpublished = {\url{https://xeno-canto.org}},
  note = {Accessed: 2026-06-22}
}

@misc{macaulay,
  author = {{Cornell Lab of Ornithology}},
  title = {Macaulay Library},
  howpublished = {\url{https://www.macaulaylibrary.org}},
  note = {Accessed: 2026-06-22}
}

@misc{birdnetpi,
  author = {McGuire, Patrick},
  title = {{BirdNET-Pi}},
  howpublished = {\url{https://github.com/mcguirepr89/BirdNET-Pi}},
  year = {2021},
  note = {Accessed: 2026-06-22}
}

@misc{iree,
  author = {{IREE}},
  title = {{Intermediate Representation Execution Environment}},
  howpublished = {\url{https://github.com/iree-org/iree}},
  note = {Accessed: 2026-06-22}
}

@article{perch,
  author = {Ghani, Burooj and Denton, Tom and Kahl, Stefan and Klinck, Holger},
  title = {Global Birdsong Embeddings Enable Superior Transfer Learning for Bioacoustic Classification},
  journal = {Scientific Reports},
  volume = {13},
  pages = {22876},
  year = {2023},
  doi = {10.1038/s41598-023-49989-z}
}

@article{audiomoth,
  author = {Hill, Andrew P. and others},
  title = {{AudioMoth}: Evaluation of a Smart Open Acoustic Device for Monitoring Biodiversity and the Environment},
  journal = {Methods in Ecology and Evolution},
  volume = {9},
  number = {5},
  pages = {1199--1211},
  year = {2018},
  doi = {10.1111/2041-210X.12955}
}

@inproceedings{tflitemicro,
  author = {David, Robert and others},
  title = {{TensorFlow Lite Micro}: Embedded Machine Learning for {TinyML} Systems},
  booktitle = {MLSys},
  year = {2021}
}

@article{cmsisnn,
  author = {Lai, Liangzhen and others},
  title = {{CMSIS-NN}: Efficient Neural Network Kernels for {Arm Cortex-M} CPUs},
  journal = {arXiv preprint arXiv:1801.06601},
  year = {2018}
}

@inproceedings{tvm,
  author = {Chen, Tianqi and others},
  title = {{TVM}: An Automated End-to-End Optimizing Compiler for Deep Learning},
  booktitle = {Proc. USENIX OSDI},
  year = {2018}
}

@inproceedings{mcunet,
  author = {Lin, Ji and others},
  title = {{MCUNet}: Tiny Deep Learning on {IoT} Devices},
  booktitle = {NeurIPS},
  year = {2020}
}

@inproceedings{utoe,
  author = {Huang, Zhaolan and others},
  title = {{U-TOE}: Universal {TinyML} On-board Evaluation Toolkit for Low-Power {IoT}},
  booktitle = {Proc. IFIP/IEEE PEMWN},
  year = {2023},
}

@article{riotml,
  author = {Huang, Zhaolan and Zandberg, Koen and Schleiser, Kaspar and Baccelli, Emmanuel},
  title = {{RIOT-ML}: Toolkit for Over-the-air Secure Updates and Performance Evaluation of {TinyML} Models},
  journal = {Annals of Telecommunications},
  year = {2025},
  doi = {10.1007/s12243-024-01041-5}
}

@inproceedings{arielml,
  author = {Huang, Zhaolan and others},
  title = {{Ariel-ML}: Computing Parallelization with Embedded {Rust} for Neural Networks on Heterogeneous Multi-core Microcontrollers},
  booktitle = {IEEE DCOSS-IoT Workshop (ISIoT)},
  year = {2026},
}

@inproceedings{msfcnn,
  author = {Huang, Zhaolan and Baccelli, Emmanuel},
  title = {{msf-CNN}: Patch-based Multi-Stage Fusion with Convolutional Neural Networks for {TinyML}},
  booktitle = {NeurIPS},
  year = {2025},
}

@misc{wrennet,
  author = {Ciapponi, Stefano and others},
  title = {Enabling Multi-Species Bird Classification on Low-Power Bioacoustic Loggers},
  year = {2025},
  note = {arXiv:2509.20103}
}

@inproceedings{disabato,
  author = {Disabato, Simone and others},
  title = {Birdsong Detection at the Edge with Deep Learning},
  booktitle = {Proc. IEEE SMARTCOMP},
  year = {2021},
  doi = {10.1109/SMARTCOMP52413.2021.00022}
}

@inproceedings{tinybird,
  author = {Schulthess, Lukas and others},
  title = {{TinyBird-ML}: An Ultra-Low Power Smart Sensor Node for Bird Vocalization Analysis and Syllable Classification},
  booktitle = {Proc. IEEE ISCAS},
  year = {2023}
}

@article{miquel,
  author = {Miquel, Jonathan and Latorre, Laurent and Chamaill{\'e}-Jammes, Simon},
  title = {Energy-Efficient Audio Processing at the Edge for Biologging Applications},
  journal = {Journal of Low Power Electronics and Applications},
  volume = {13},
  number = {2},
  pages = {30},
  year = {2023},
  doi = {10.3390/jlpea13020030}
}

@misc{unpu,
  author = {Millar, Josh and others},
  title = {Benchmarking Ultra-Low-Power {$\mu$}{NPUs}},
  year = {2025},
  note = {arXiv:2503.22567}
}

@misc{max78000,
  author = {{Analog Devices}},
  title = {{MAX78000}},
  howpublished = {\url{https://www.analog.com/en/products/max78000.html}},
  note = {Accessed: 2026-06-22}
}

@misc{ethosu55,
  author = {{Arm}},
  title = {{Arm Ethos-U55}: {microNPU} for Embedded Machine Learning},
  howpublished = {\url{https://www.arm.com/products/silicon-ip-cpu/ethos/ethos-u55}},
  note = {Accessed: 2026-06-22}
}

@misc{stm32n6,
  author = {{STMicroelectronics}},
  title = {{STM32N6} Series},
  howpublished = {\url{https://www.st.com/en/microcontrollers-microprocessors/stm32n6-series.html}},
  note = {Accessed: 2026-06-22}
}

@misc{rfc7228,
  author = {Bormann, Carsten and Ersue, Mehmet and Ker{\"a}nen, Ari},
  title = {{Terminology for Constrained-Node Networks}},
  howpublished = {RFC 7228},
  publisher = {RFC Editor},
  year = {2014},
  doi = {10.17487/RFC7228}
}

@article{abdoli,
  author = {Abdoli, Sajjad and Cardinal, Patrick and Lameiras Koerich, Alessandro},
  title = {End-to-End Environmental Sound Classification using a {1D} Convolutional Neural Network},
  journal = {Expert Systems with Applications},
  volume = {136},
  pages = {252--263},
  year = {2019},
  doi = {10.1016/j.eswa.2019.06.040}
}

@inproceedings{sang,
  author = {Sang, Jonghee and others},
  title = {Convolutional Recurrent Neural Networks for Urban Sound Classification Using Raw Waveforms},
  booktitle = {Proc. EUSIPCO},
  year = {2018},
  doi = {10.23919/EUSIPCO.2018.8553247}
}

@misc{kws,
  author = {Wang, Jingyi and Li, Shengchen},
  title = {Keyword Spotting System and Evaluation of Pruning and Quantization Methods on Low-Power Edge Microcontrollers},
  year = {2022},
  note = {arXiv:2208.02765}
}

@misc{mlperftiny,
  author = {Banbury, Colby and others},
  title = {{MLPerf} Tiny Benchmark},
  year = {2021},
  note = {arXiv:2106.07597}
}

@article{thapa2026embedded-rust,
  title={{Embedded Rust or C Firmware? Lessons from an Industrial Microcontroller Use Case with Ariel OS}},
  author={Thapa, Bipin and others},
  journal={IEEE DCOSS-IoT Workshop (IoTI5)},
  year={2026}
}

\end{document}